\documentclass[conference]{IEEEtran}
\IEEEoverridecommandlockouts
\usepackage{cite}
\usepackage{amsmath,amssymb,amsfonts}
\usepackage{algorithmic}
\usepackage{algorithm}
\usepackage{caption} 
\usepackage{graphicx}
\usepackage{textcomp}
\usepackage{xcolor}
\usepackage{multirow}
\def\BibTeX{{\rm B\kern-.05em{\sc i\kern-.025em b}\kern-.08em
    T\kern-.1667em\lower.7ex\hbox{E}\kern-.125emX}}
\begin{document}

\title{GC-ConsFlow: Leveraging Optical Flow Residuals and Global Context for Robust Deepfake Detection}

\author{
    \IEEEauthorblockN{
        Jiaxin Chen\textsuperscript{1}, 
        Miao Hu\textsuperscript{1}, 
        Dengyong Zhang\textsuperscript{1,2\textasteriskcentered}, 
        Jingyang Meng\textsuperscript{1}
    }
    \IEEEauthorblockA{\textsuperscript{1}Changsha University of Science and Technology, Changsha, China \\
    }
}

\maketitle

\begin{abstract}
The rapid development of Deepfake technology has enabled the generation of highly realistic manipulated videos, posing severe social and ethical challenges. Existing Deepfake detection methods primarily focused on either spatial or temporal inconsistencies, often neglecting the interplay between the two or suffering from interference caused by natural facial motions. To address these challenges, we propose the global context consistency flow (GC-ConsFlow), a novel dual-stream framework that effectively integrates spatial and temporal features for robust Deepfake detection. The global grouped context aggregation module (GGCA), integrated into the global context-aware frame flow stream (GCAF), enhances spatial feature extraction by aggregating grouped global context information, enabling the detection of subtle, spatial artifacts within frames. The flow-gradient temporal consistency stream (FGTC), rather than directly modeling the residuals, it is used to improve the robustness of temporal feature extraction against the inconsistency introduced by unnatural facial motion using optical flow residuals and gradient-based features. By combining these two streams, GC-ConsFlow demonstrates the effectiveness and robustness in capturing complementary spatiotemporal forgery traces. Extensive experiments show that GC-ConsFlow outperforms existing state-of-the-art methods in detecting Deepfake videos under various compression scenarios.
\end{abstract}


\section{Introduction}
\label{sec:intro}

Recently, advancements in deep learning and computer vision have enabled the generation of highly realistic manipulated videos, such as those created by replacing actions \cite{b1} and altering facial expressions \cite{b2}. While Deepfake technology has positive applications in entertainment and art, its misuse has caused significant societal challenges \cite{b3}. For example, a fabricated video of ``AI Lee Ka-chiu" promoting a ``high-return investment plan" went viral in January 2024, creating public panic in Hong Kong. This highlights the urgent need for effective Deepfake detection mechanisms.

Existing frame-based detection methods \cite{b4, b5} primarily analyzed individual frames to identify forgeries but failed to capture temporal inconsistencies unique to Deepfake videos. Recent studies \cite{b7, b8, b9} incorporated temporal and spatial information to improve detection but often neglected interference from natural facial motion, including expression changes or head movements, leading to suboptimal performance.

To address these limitations, we propose a global context consistency flow-based network (GC-ConsFlow), which extracts both frame-level and temporal-level features through the global context-aware frame flow stream (GCAF) and the flow-gradient temporal consistency stream (FGTC). Specifically, the GCAF stream captures global artifact features within frames using the proposed global grouped context aggregation module (GGCA), which extracts multidimensional global information by performing global average pooling and max pooling along the height and width dimensions of the feature map and generates attention weights to enhance the feature representation capability of global artifact features. The FGTC stream introduces a novel concept in video forensics by utilizing optical flow residuals to extract artifacts unrelated to natural facial movements. This stream effectively distinguish natural facial motion from forgery-induced anomalies and improve detection accuracy. Gradient features are also introduced as complementary inputs to better capture frame-to-frame variations. The main contributions of this study are as follows:
\begin{itemize}
  \item We introduce the concept of optical flow residuals into Deepfake detection, offering a novel perspective. Unlike existing methods that rely on pixel-level inconsistencies, GC-ConsFlow leverages motion-compensated optical flow residuals combined with gradient-based complementary features to effectively suppress interference from natural facial movements, significantly enhancing the detection of forgery artifacts.
  \item We propose a novel Deepfake detection network that incorporates specialized modules to improve feature extraction capabilities. The GGCA module in the GCAF stream aggregates global contextual information along the height and width dimensions, enhancing the spatial feature representation and enabling more effective detection of subtle forgery traces in videos.
  \item We conduct several experiments to evaluate the effectiveness of our GC-ConsFlow. Experimental results demonstrate that GC-ConsFlow achieves superior performance compared to state-of-the-art methods on public datasets, which is robust to heavy compression.
\end{itemize}

\begin{figure*}[htbp]
\centering
\includegraphics[width=0.98\textwidth]{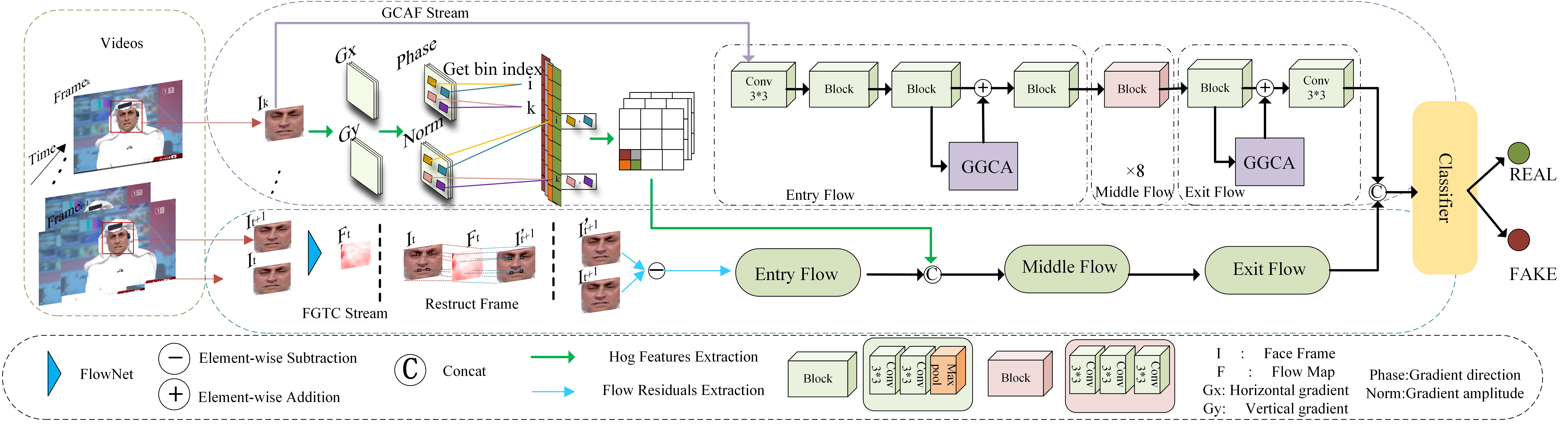}
\caption{Overview of the proposed GC-ConsFlow framework for deepfake detection.}
\label{Fig1}
\end{figure*}

\section{METHODOLOGY}
Fig. \ref{Fig1} illustrate the proposed GC-ConsFlow network, which consists of GCAF stream and FGTC stream. The GCAF stream is designed based on the principle of global contextual modeling, aiming to detect subtle spatial artifacts left by Deepfake generation. The FGTC stream leverages optical flow residuals and gradient-based features to model anomalous artifacts in non-facial natural motion within forged videos. Each stream is elaborated in detail below.

\subsection{Global Context-Aware Frame Flow Stream}
Existing DeepFakes generation techniques typically create forged videos by blending the target and source faces. Although advancements in generation methods have significantly improved face quality and post-processing techniques have masked many visible artifacts, certain unnatural spatial artifacts still persist in facial fusion regions. These artifact features are difficult to eliminate completely and can be effectively captured by deep convolutional neural networks.

Based on the above analysis, we designed the GCAF stream to effectively detect tampering traces in DeepFake videos by focusing on spatial artifact features within individual frames. Specifically, the input consists of \(T\) cropped and rotated RGB frames of size \(B \times C \times H \times W\), where \(B\) is the batch size, \(C\) is the number of channels, and \(H\) and \(W\) are the frame height and width. These frames are fed into XceptionNet \cite{chollet2017xception} to extract spatial features, which serves as the backbone for the GCAF stream. To further enhance the spatial features, we introduce the GGCA module. By applying multidimensional global pooling along the height and width dimensions, the GGCA module effectively captures global forgery features and improves feature representation. It is integrated into both the Entry flow and Exit flow of the XceptionNet, enabling robust spatial feature extraction.
\begin{figure}
\includegraphics[width=0.49\textwidth]{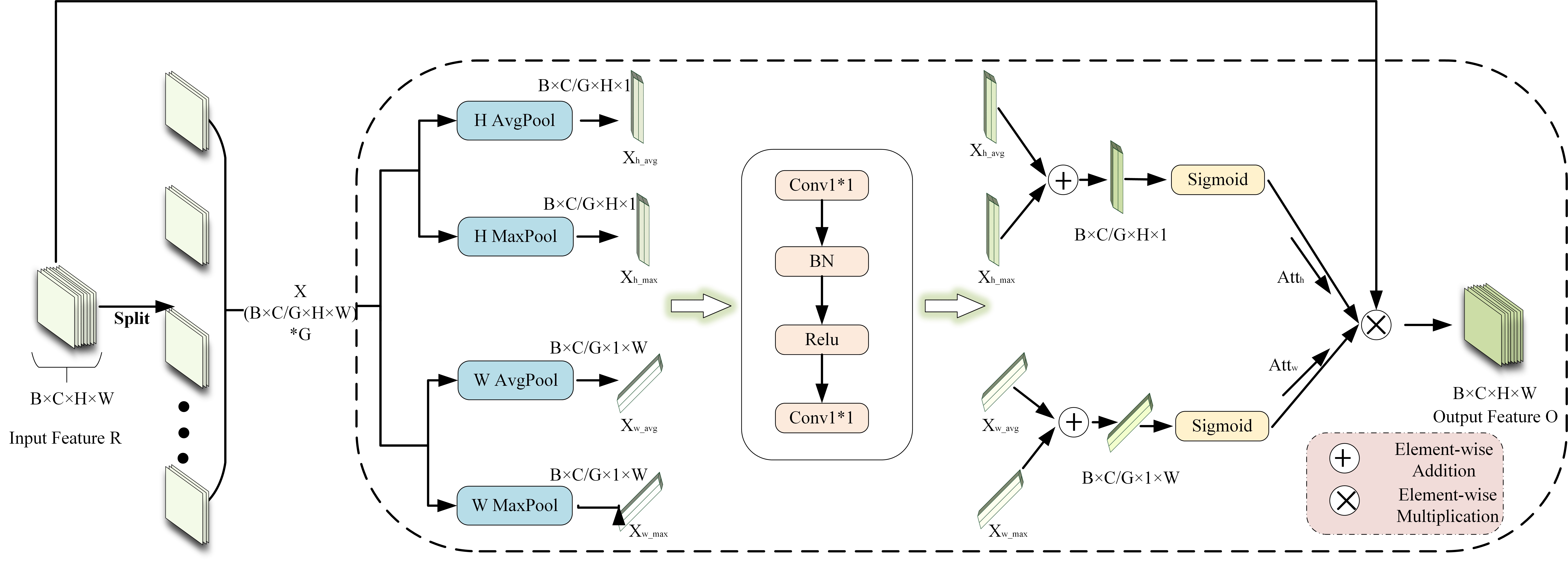}	
\caption{The Global Grouped Context Aggregation (GGCA) is designed to enhance spatial feature extraction by aggregating grouped global context information.}
\label{Fig2}
\end{figure}
As shown in Fig. \ref{Fig2}, the GGCA module processes the input feature \( R \ (B \times C \times H \times W) \) by dividing it into \( G \) groups along the channel dimension, effectively reducing the computational cost for each group. Each group contains \( C/G \) channels, resulting in the feature \( X \ (B \times C/G \times H \times W) \). To extract global features from multiple dimensions, average pooling is employed to capture overall distribution trends, while max pooling preserves locally salient features. The combination of these two pooling methods enables a more comprehensive extraction of forgery traces. Additionally, operations along the height and width dimensions generate multi-directional global information, denoted as \( X_{h\_max}, X_{h\_avg}, X_{w\_max}, X_{w\_avg} \), 
further enhancing the completeness of feature extraction. This process can be expressed using the following formulas.
\begin{equation}
X_{h_{\_avg}}=AvgPool(X) ,X_{h_\_max}=MaxPool(X)\\
\end{equation}
\begin{equation}
X_{w_{\_avg}}=AvgPool(X),X_{w_\_max }=MaxPool(X)\\
\end{equation}

Subsequently, a 1×1 convolution is applied to the four feature maps to restore the channel dimension. The resulting feature maps are summed along the height and width dimensions, and the sigmoid function is used to generate height-oriented and width-oriented attention weights, $Att_{h}$ and $Att_{w}$.
\begin{equation}
Att_{h}=Sigmoid(X_{h_{\_avg}}+X_{h_\_max})\\
\end{equation}
\begin{equation}
Att_{w}=Sigmoid(X_{w_{\_avg}}+X_{w_\_max})\\
\end{equation}

Finally, the input feature \( R \ (B \times C \times H \times W) \) is weighted according to the attention weights in the height and width directions, resulting in the refined feature \( O \ (B \times C \times H \times W) \), which enhances the representation of key spatial regions.
\begin{equation}
O=R*Att_{h}*Att_{w}\\
\end{equation}
The spatial features O of the T images are summed and averaged to serve as the final output of the GCAF stream.

Although the GCAF stream can learn spatial artifact features within individual frames, it focuses solely on spatial features of single frames and cannot capture temporal information between frames. 

\subsection{Flow-Gradient Temporal Consistency Stream}
Based on the previous analysis, we know that one key difference between real and fake videos is that fake videos often ignore temporal consistency during the synthesis process. Since our GCAF stream primarily focuses on facial regions within individual frames, the proposed FGTC stream should instead focus on relationships between frames. Additionally, residuals have been proven to be temporal-related features. Therefore, the FGTC stream leverages temporal-related residual features to extract inconsistencies between frames, while also incorporating gradient-based complementary features. Specifically, gradient features, represented by histograms of oriented gradients (HOG), describe texture and edge information within frames, serving as auxiliary features for modeling temporal characteristics. By combining optical flow residuals with gradient features, we use XceptionNet for classification, enabling efficient detection of video forgeries. The details are as follows.

Compared with traditional residual feature extraction methods in video forgery detection, our approach introduces the concept of optical flow residuals for temporal inconsistency modeling, aiming to improve the ability to capture abnormal dynamic changes between frames. Traditional video forgery detection methods typically calculate pixel-level residuals between consecutive video frames as features to detect inter-frame inconsistencies. However, this approach suffers from a significant drawback: due to natural facial movements introducing complex pixel-level variations, directly using pixel residuals is prone to interference from these natural motions, resulting in a large amount of unnecessary noise and reducing the accuracy of forgery detection.

To address the issue where traditional pixel-level residual features are prone to interference from complex variations caused by natural facial movements, leading to excessive noise and reduced accuracy in forgery detection, our method differs from traditional residual feature extraction methods by fully leveraging the advantages of optical flow in capturing motion information of objects within a scene. Optical flow effectively describes local motion patterns between frames and is less susceptible to pixel-level noise. Therefore, when extracting temporal inconsistency features, we introduce the concept of optical flow residuals, analyzing them to capture the lack of temporal consistency in forged videos.

We consider two consecutive frames, \(I_t\) and \(I_{t+1}\). First, we use the FlowNet\cite{b26} to compute the optical flow map \(F_t\) between \(I_t\) and \(I_{t+1}\). Subsequently, we utilize the optical flow feature \(F_t\) and frame \(I_t\) to reconstruct the motion-compensated next frame, \(I_{t+1}^{'}\). Specifically, for each pixel in the reconstructed frame \(I_{t+1}^{'}\), its corresponding position in frame \(I_t\) is determined by a normalized grid coordinate \(G[h, w]\), where \(h\) denotes the horizontal coordinate and \(w\) denotes the vertical coordinate.
\begin{equation}
G(h,w)=[\frac{2w}{W-1},\frac{2h}{H-1}]\\
\end{equation}
The optical flow \(F_t\) is added to the \(G[h, w]\) to obtain the pixel coordinates \({G}^{’}(h, w)\) for \(I_{t+1}^I\). To ensure the pixel coordinates do not exceed the image boundaries, a clamp operation is applied to restrict \({G}^{'}(h, w)\) to the range \([-1, 1]\). This process can be expressed by the following equation.  
\begin{equation}
{G}^{'}(h,w)=clamp[G(h,w)+\frac{F_{t}(h,w)}{W,H},(-1,1)]\\
\end{equation}
The target pixel values of \(I_{t+1}^{’}\) are sampled from \(I_t\) based on the target coordinates \({G}^{'}(h, w)\). Each component of \({G}^{'}(h, w)\) can be decomposed as \({G}^{’}(h, w) = [{X}^{'}, {Y}^{’}] \in (B \times H \times W \times 2)\), where \(({X}^{'}, {Y}^{'}) \in [-1, 1]\) represents the normalized target coordinates. To reconstruct the pixel values of \(I_{t+1}^{'}\), the target coordinates are mapped back to the original pixel coordinate system to obtain \((X_{\text{real}}, Y_{\text{real}})\). Using these coordinates, the pixel values are interpolated as a weighted sum of neighboring pixels.
\begin{equation}
X_{real}=\frac{(X^{'}+1)*(W-1)}{2},Y_{real}=\frac{(Y^{'}+1)*(H-1)}{2}\\
\end{equation}
The integer parts of \(X_{\text{real}}\) and \(Y_{\text{real}}\) represent the coordinates of the top-left corner of the neighboring pixel, denoted as \((\lfloor X_{real} \rfloor,\lfloor Y_{real} \rfloor)\). From this, four neighboring pixel coordinates can be derived: \((\lfloor X_{real} \rfloor,\lfloor Y_{real} \rfloor)\), \((\lfloor X_{real} \rfloor+1,\lfloor Y_{real} \rfloor)\), \((\lfloor X_{real} \rfloor,\lfloor Y_{real} \rfloor+1)\), and \((\lfloor X_{real} \rfloor+1,\lfloor Y_{real} \rfloor+1)\). The fractional parts of \(X_{\text{real}}\) and \(Y_{\text{real}}\) represent the coordinate offsets, which are used to calculate the interpolation weights for each pixel.  
\begin{equation}
\begin{aligned}
& W_{00}=[1-(X_{real}-\lfloor X_{real} \rfloor)][1-(Y_{real}-\lfloor Y_{real} \rfloor)],\\
& W_{10}=(X_{real}-\lfloor X_{real} \rfloor)[1-(Y_{real}-\lfloor Y_{real} \rfloor)],\\
& W_{01}=[1-(X_{real}-\lfloor X_{real} \rfloor)](Y_{real}-\lfloor Y_{real} \rfloor),\\
&W_{11}=(X_{real}-\lfloor X_{real} \rfloor)(Y_{real}-\lfloor Y_{real} \rfloor)
\end{aligned}
\end{equation}
The weighted sum of the neighboring pixels is computed to obtain the pixel value at the corresponding relative position in the reconstructed frame \(I_{t+1}^{'}\), thereby generating the reconstructed frame \(I_{t+1}^{'}\). 
\begin{equation}
I_{t+1}^{'}=\{\sum_{i=0}^1\sum_{j=0}^1W_{ij}\cdot I_{t}(x,y) \mid x\in[0,W-1],y\in[0,H-1]\}
\end{equation}
We hypothesize that the reconstructed frame \(I_{t+1}^{'}\) better reflects the motion pattern of the current frame \(I_t\). The residual between the reconstructed frame \(I_{t+1}^{'}\) and the original frame \(I_{t+1}\), denoted as \(Residual\), captures the motion-compensated inconsistencies. These inconsistencies may indicate anomalous motions in forged regions rather than natural variations caused by the normal motion of objects. This hypothesis is validated in the experimental section.
\begin{equation}
Residual=Abs(I_{t+1}^{'}-I_{t+1})
\end{equation}

However, merely relying on residual features may not be sufficient, as residuals only capture the inter-frame differences after motion compensation. For certain more sophisticated video forgery techniques, such differences might be concealed or weakened. Since gradient features are robust to low-frequency variations induced by illumination and compression, to more comprehensively characterize the dynamic relationships between video frames, we introduce gradient features for continuous frames\cite{b27} as a complement.
\begin{algorithm}
\caption{HOG Feature Extraction}
\renewcommand{\algorithmicrequire}{\textbf{Input:}}
\renewcommand{\algorithmicensure}{\textbf{Output:}}
\begin{algorithmic}[1]
\REQUIRE  $x \in \mathbb{R}^{B \times 3 \times H \times W}$, $pool$: pooling size\\
\ENSURE   $Hog_{out} \in \mathbb{R}^{B \times 3 \times 9 \times \frac{H}{pool} \times \frac{W}{pool}}$   

\STATE \textbf{Initialize:} Sobel filters $weight_x$, $weight_y$

\STATE Pad $x$ with reflection padding
\STATE Compute gradients: 
\[
gx = \text{conv2d}(x, weight_x), \quad gy = \text{conv2d}(x, weight_y)
\]
\STATE Compute gradient magnitude and orientation:
\[
\text{norm} = \sqrt{gx^2 + gy^2}, \quad \text{phase} = \arctan(\frac{gy}{gx})
\]

\STATE \textbf{Initialize} histogram $out = \text{zeros}(B, 3, 9, H, W)$

\STATE Scatter-add gradient magnitudes into orientation bin:
\[
\text{out}\left[\left\lfloor \frac{\text{phase} + \pi}{2\pi} \cdot 9 \right\rfloor\right] += \text{norm}
\]

\STATE Image chunking and accumulating histograms of gradient directions within blocks:
\[
Hog_{out} = \sum_{\text{pool \(\times\) pool} \in \text{H \(\times\) W}} out_{\text{pool \(\times\) pool}}
\]

\STATE L2 Normalize histograms along orientation axis.
\RETURN $Hog_{out}$
\end{algorithmic}
\end{algorithm}
Specifically,the horizontal gradient \(g_{x}\) and vertical gradient \(g_{y}\) of the image are first computed using the weight matrices \(weight_{x}\) and \(weight_{y}\). 
\begin{equation}
	\begin{gathered}
	 weight_{x}=\begin{bmatrix}1 & 0 & -1 \\2 & 0 & -2 \\1 & 0 & -1\end{bmatrix}	
	\end{gathered}
    ,weight_{y}=weight_{x}^T
\end{equation}

Based on \(g_{x}\) and \(g_{y}\), the gradient magnitude \(norm\) and gradient direction \(phase\) are calculated for each pixel. For each pixel, the  \(phase\) is mapped to the \([0, 9]\) interval to determine the corresponding bin index in the gradient orientation histogram. The \(norm\) of each pixel is then used to perform weighted accumulation for the corresponding bin in the orientation histogram. Next, the image is divided into grid blocks of size pool × pool, where the gradient orientation histograms within each block are aggregated. Finally, to enhance robustness against illumination variations, L2 normalization is applied to the gradient orientation histograms of each block, generating the final histogram of oriented gradient (HOG) feature map $Hog_{out}$ for the image.The detailed algorithm is provided in Algorithm 1.

We first feed the motion-compensated residual features into the shallow convolutional layers of XceptionNet for feature extraction. Subsequently, these features are concatenated with the gradient features of consecutive video frames and then input into the remaining layers of XceptionNet to detect inconsistencies between frames in compressed Deepfake videos. Finally, the outputs of the frame-level stream and the temporal stream are concatenated and passed through a sigmoid function to obtain the final video-level classification result of the video.

\section{EXPERIMENTS}
\subsection{ Experimental Settings}
\textbf{Datasets: }For the experiments in this study, we conducted experiments using two widely-used datasets in the field of facial forgery detection: FaceForensics++ (FF++) \cite{b28} and Celeb-DF \cite{b29}. The FF++ dataset contains four types of manipulated videos generated by different forgery techniques: DeepFakes(DF), Face2Face(F2F), FaceSwap(FS), and NeuralTextures(NT), resulting in a total of 1,000 forged videos for each method (4,000 in total). Additionally, FF++ provides three versions of videos with varying levels of compression: raw (uncompressed), C23(lightly compressed, HQ), and C40(heavily compressed, LQ). For our experiments, we focus on the HQ and LQ subsets. Specifically, we use 80\% of all videos for training and the remaining 20\% for testing. Furthermore, we extract the first 200 frames from each video to conduct our experiments. Celeb-DF, containing 5,639 high-quality DeepFake videos with realistic details, is used for model generalization testing. 

\textbf{Evaluation Metrics: }In this experiment, we use Accuracy (ACC) and Area Under Curve(AUC) as our measurement indicators to validate DeepFake detection performance. 

\textbf{Implemental Details: } We implement the proposed method using the PyTorch framework and train it on an NVIDIA 3060Ti GPU. The input sequence consists of 6 frames, each with 3 channels and a resolution of 224×224. The network is trained end-to-end using the Adam optimizer with binary cross-entropy as the loss function. The learning rate is set to 0.0001, and the batch size is 6. The model is trained for 30 epochs, with a total training time of approximately 12 hours.

\begin{table}[]
\renewcommand{\arraystretch}{1.2} 
\setlength{\tabcolsep}{4pt} 
\captionsetup{justification=centering, singlelinecheck=false} 
\caption{ Ablation research of several modules on FF++ (LQ). The evaluation metric is ACC.}

\begin{tabular}{ccllll}
\hline
\multicolumn{1}{l}{} & \textbf{Models}                         & \multicolumn{1}{c}{\textbf{DF}}      & \multicolumn{1}{c}{\textbf{F2F}}     & \multicolumn{1}{c}{\textbf{FS}}      & \multicolumn{1}{c}{\textbf{NT}}      \\ \hline
\textbf{1}           & \textbf{GCAF Stream}                    & 94.04\%                              & 86.79\%                              & 92.37\%                              & 76.16\%                              \\
\textbf{2}           & \textbf{FGTC Stream}                    & 86.86\%                              & 76.29\%                              & 85.01\%                              & 66.76\%                              \\
\textbf{3}           & \textbf{w/o GGCA}           & 94.54\%                              & 86.90\%                              & 93.09\%                              & 77.20\%                              \\

\textbf{4}           & \textbf{w/o Reconstruct Frame}          & 83.05\%                              & 74.10\%                              & 82.04\%                              & 62.90\%   \\             
\textbf{5}           & \textbf{Reconstruct Frame} & 86.40\%                              & 75.50\%                              & 84.60\%                              & 63.78\%                              \\
\textbf{6}           & \textbf{GC-ConsFlow}                    & \multicolumn{1}{c}{\textbf{94.82\%}} & \multicolumn{1}{c}{\textbf{87.21\%}} & \multicolumn{1}{c}{\textbf{93.83\%}} & \multicolumn{1}{c}{\textbf{78.49\%}} 
\\ \hline
\label{ablation}
\end{tabular}
\end{table}
\subsection{Ablation Study}
\textbf{Evaluation of Single Stream:} To evaluate the performance of each stream, we compare our GCAF stream, FGTC stream with our proposed dual-stream network. As shown in Table \ref{ablation}, the detection accuracy of the GCAF stream (row 1) and the FGTC stream (row 2) is lower than that of our GC-ConsFlow method (row 6). This indicates that combining both streams enhances Deepfake video detection performance. This is because the GCAF stream focuses on capturing spatial features within individual frames, effectively identifying spatial artifact features. On the other hand, the FGTC stream is designed to model unnatural motion patterns caused by forgery. While each stream independently captures specific aspects of forgery traces (spatial or temporal), they are limited in understanding the interplay between spatial artifacts and temporal inconsistencies. Spatial artifacts may only become evident in the context of temporal changes, while temporal anomalies often appear in regions affected by spatial manipulations. Ignoring this relationship will cause incomplete forgery detection. By combining the two streams, GC-ConsFlow integrates spatial and temporal information, leveraging the interdependence between frame-level and inter-frame features. This complementary fusion enhances overall feature representation, enabling the model to more comprehensively capture spatiotemporal forgery traces and significantly improve detection performance.

\textbf{Evaluation of GGCA Module:} To validate the effectiveness of the GGCA module, we compare network without using the GGCA module (w/o GGCA) with our GC-ConsFlow network, as shown in Table \ref{ablation} (row 3 and row 6). Network without the GGCA module performed at least 0.28\% worse, demonstrating the positive impact of the GGCA module. This improvement stems from the GGCA module's ability to enhance spatial feature by aggregating grouped global context information. By generating attention weights along spatial dimensions and applying them to input features, the module emphasizes key regions, improving feature representation for detecting subtle artifacts. This enhanced spatial modeling directly boosts the performance of the GC-ConsFlow framework.

\begin{figure}
\centering
\includegraphics[width=0.29\textwidth]{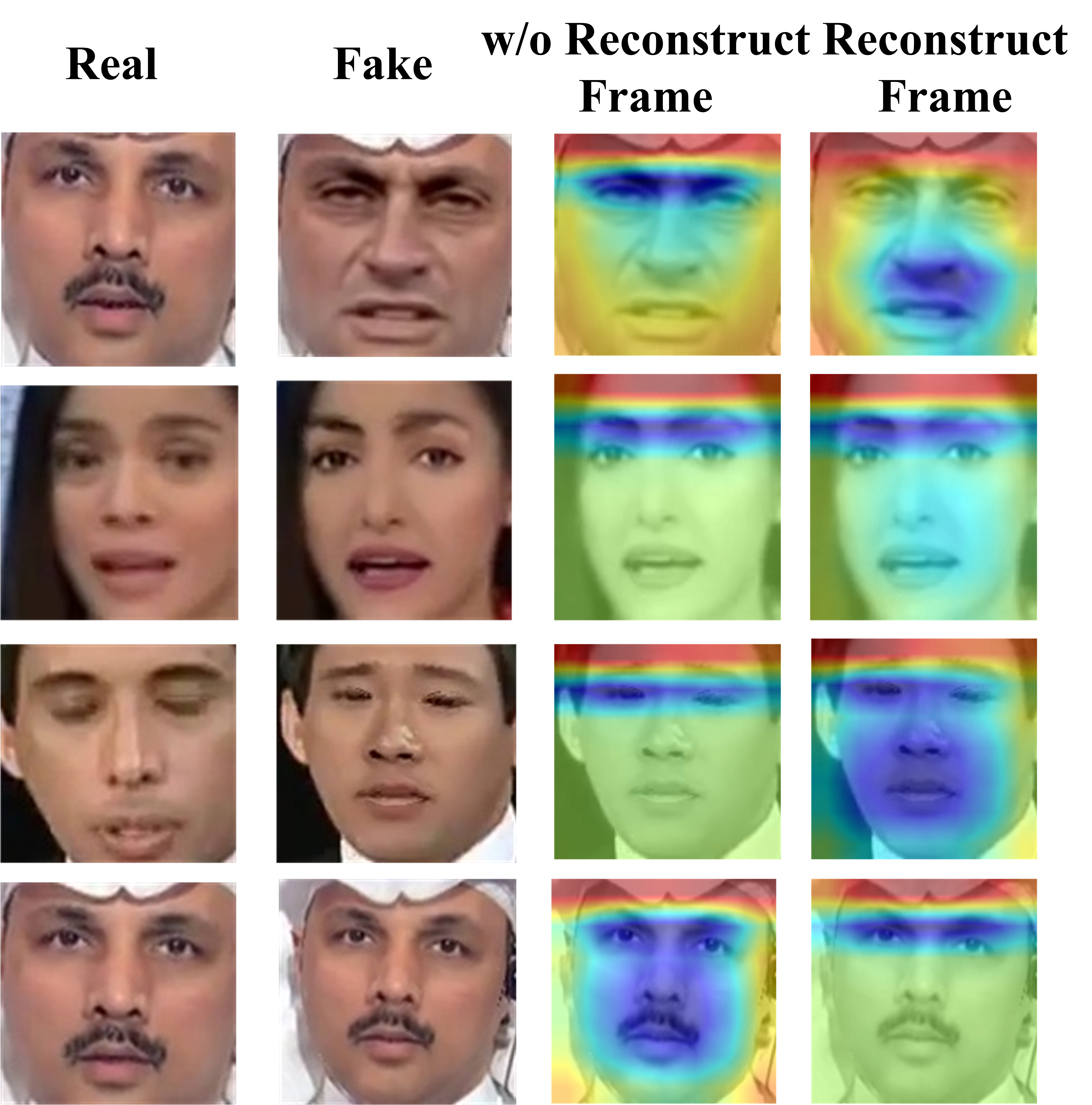}	
\caption{Comparison of heatmaps generated with and without reconstructed frames on real and forged video frames.}
\label{Fig3}
\end{figure}

\textbf{Evaluation of Reconstructed Frame:} To prove the hypothesis that the reconstructed frame \(I_{t+1}^{'}\) better aligns with the motion trajectory of frame \(I_{t}\), we conduct comparison experiments detailed in rows 4 and 5 of Table \ref{ablation}. To isolate the impact of gradient features on residual features, row 4 computes residuals between adjacent original frames, while row 5 computes residuals between adjacent reconstructed frames. These residuals are then fed into XceptionNet without additional features. As shown in Table \ref{ablation}, the residuals between adjacent reconstructed frames consistently outperform those between original frames across all four FF++ LQ subsets. 

Fig. \ref{Fig3} further illustrates this difference. The first column shows real video frames, the second column forged frames, the third column heatmaps without reconstructed frames, and the fourth column heatmaps with reconstructed frames. In the first three rows, where entire faces are manipulated, the model with reconstructed frames highlights the entire face, while the model without them focuses only on localized areas, such as differences around the eyes in the second row. In the fourth row, where only the eyes are manipulated, the model with reconstructed frames effectively isolates unnatural motion artifacts in the manipulated region, whereas the model without reconstructed frames is distracted by noise from natural facial movements. This evidence supports our hypothesis. Additionally, experiments in rows 2 and 5 of Table \ref{ablation} show that incorporating HOG features significantly improves model performance, further confirming their effectiveness.

\begin{table*}[]
\centering
\renewcommand{\arraystretch}{1.2} 
\setlength{\tabcolsep}{8pt} 
\captionsetup{justification=centering, singlelinecheck=false} 
\caption{Performance comparison on the FF++ dataset under HQ and LQ compression levels, as well as generalization evaluation on the Celeb-DF dataset. Experimental data from \cite{b41} are sourced from the corresponding paper. Best results are bolded, with some reproduced (*) and others taken from \cite{b49}. Metrics: ACC (FF++) and AUC (Celeb-DF).}

\begin{tabular}{cllllllllc}
\hline
\multirow{2}{*}{\textbf{Methods}}              & \multicolumn{4}{c}{\textbf{FF++ HQ}}                                                                                                                  & \multicolumn{4}{c}{\textbf{FF++ LQ}}                                                                                                                  & \multirow{2}{*}{\textbf{Celeb-DF}} \\ \cline{2-9}
                                               & \multicolumn{1}{c}{\textbf{DF}} & \multicolumn{1}{c}{\textbf{F2F}} & \multicolumn{1}{c}{\textbf{FS}} & \multicolumn{1}{c}{\textbf{NT}} & \multicolumn{1}{c}{\textbf{DF}} & \multicolumn{1}{c}{\textbf{F2F}} & \multicolumn{1}{c}{\textbf{FS}} & \multicolumn{1}{c}{\textbf{NT}} &                                                   \\ \hline
\textbf{C3D\cite{b32}}                          & 92.86\%                             & 88.57\%                              & 91.97\%                             & 89.64\%                             & 89.86\%                             & 82.86\%                              & 87.86\%                             & 87.14\%                             & \textbf{-}                         \\
\textbf{I3D\cite{b33}}                          & 92.80\%                             & 92.86\%                              & 96.43\%                             & 90.36\%                             & 91.07\%                             & 86.43\%                              & 91.43\%                             & 78.57\%                             & 74.11\%                            \\
\textbf{ADDNet-3D \cite{b37}}                    & 92.14\%                             & 83.93\%                              & 92.50\%                             & 78.21\%                             & 90.36\%                             & 78.21\%                              & 80.00\%                             & 69.29\%                             & 60.85\%                            \\
\textbf{MSVT  \cite{b38}}                         & 95.79\%                             & 93.72\%                              & 92.93\%                             & 92.24\%                             & \multicolumn{1}{c}{\textbf{-}}      & \multicolumn{1}{c}{\textbf{-}}       & \multicolumn{1}{c}{\textbf{-}}      & \multicolumn{1}{c}{\textbf{-}}      & \textbf{-}                         \\
\textbf{FAMM* \cite{b39}}                         & 97.23\%                             &92.51\%                              & 92.68\%                             & 85.98\%         & 87.86\%                             & 87.10\%                              & 85.36\%                             & 74.29\%                             & 59.06\%                         \\
\textbf{DDLmodel* \cite{b11}}                     & 98.75\%                             & 97.57\%                              & 98.16\%                             & 88.87\%                             & 92.27\%                             & 82.01\%                              & 87.13\%                             & 70.88\%                             & 72.55\%                            \\
\textbf{LIPINC* \cite{lip}}                     & 89.71\%                             & 76.58\%                              & 84.82\%                             & 78.49\%                             & 77.78\%                             & 69.43\%                              & 65.29\%                             & 67.64\%                             & 62.75\%                            \\

\textbf{t-SFL\cite{b41}}                        & \multicolumn{1}{c}{\textbf{-}}      & \multicolumn{1}{c}{\textbf{-}}       & \multicolumn{1}{c}{\textbf{-}}      & \multicolumn{1}{c}{\textbf{-}}      & 93.90\%                             & \textbf{96.20\%}                     & 93.60\%                             & \textbf{89.00\%}                    & \textbf{-}                         \\
\multicolumn{1}{l}{\textbf{GC-ConsFlow(ours)}} & \textbf{99.00\%}                    & \textbf{98.93\%}                     & \textbf{98.80\%}                    & \textbf{93.82\%}                    & \textbf{94.82\%}                    & 87.21\%                              & \textbf{93.83\%}                    & 78.49\%                             & \textbf{75.91\%}                   \\ \hline
\end{tabular}
\label{tab:results}
\end{table*}

\subsection{Comparison to SOTA methods }
Using video-level accuracy as the evaluation metric, we compared our model with the state-of-the-art methods on the FF++ dataset under varying compression levels (Table \ref{tab:results}). On the HQ subset, GC-ConsFlow achieves the best performance across all categories, showcasing its ability to capture both spatial features and spatiotemporal inconsistencies. On the LQ subset, while GC-ConsFlow performs slightly worse than t-SFL\cite{b41} on the Face2Face and NeuralTextures subsets, it still outperforms most existing methods and delivers competitive results overall.

The reduced performance on the Face2Face subset can be attributed to forgery traces being concentrated in localized regions, such as the mouth and eyes. While the GGCA module enhances spatial feature extraction through global pooling techniques like max pooling, it may fail to fully preserve fine-grained local details in highly compressed videos, making it harder to detect subtle localized artifacts. Similarly, the NeuralTextures subset excels at producing consistent facial textures, where forgery traces are faint and further obscured by heavy compression. This compression smooths out spatiotemporal inconsistencies, diminishing the effectiveness of motion-compensated residuals in highlighting detectable differences. As a result, these subtle features are often overlooked during training, leading to reduced detection performance. Despite these challenges, our method achieves strong results on other subsets. By combining spatial features with spatiotemporal inconsistencies, GC-ConsFlow effectively captures complementary manipulation traces, demonstrating superior performance on the DeepFakes and FaceSwap datasets.

To assess the generalization ability of our model, we trained it on the HQ subset of the FF++ dataset and tested it on the Celeb-DF test set using the AUC metric (Table \ref{tab:results}). GC-ConsFlow achieved an AUC of 75.91\%, surpassing methods like DDLmodel (72.55\%) and I3D (74.11\%), demonstrating its robustness in detecting unseen forgery patterns. This strong generalization is attributed to its dual-stream design: the GCAF stream captures global spatial artifacts, while the FGTC stream models spatiotemporal inconsistencies, enabling effective adaptation to diverse forgery techniques.

\section{CONCLUSION}
This paper proposes a global context consistency flow-based network (GC-ConsFlow) that combines frame-level and temporal-level features for Deepfake detection. The GCAF stream, enhanced by the GGCA module, captures global context artifacts, while the FGTC stream uses optical flow residuals and gradient features to suppress noise caused by natural facial motion. By integrating the outputs of both streams, our GC-ConsFlow can effectively capture complementary spatiotemporal forgery traces. Experimental results show that GC-ConsFlow outperforms existing methods on multiple datasets, achieving robust performance even under heavy compression. The GGCA module and the proposed optical flow residuals significantly enhance spatial and temporal modeling, respectively, demonstrating the robustness and generalization ability of our GC-ConsFlow.

\bibliographystyle{IEEEbib}
\bibliography{icme2025}

\end{document}